\documentclass[runningheads]{llncs}
\usepackage{graphicx}

\usepackage{tikz}
\usepackage{comment}
\usepackage{amsmath,amssymb} 
\usepackage{color}
\usepackage{wrapfig}

\usepackage[accsupp]{axessibility}  
\usepackage{tabularx}
\usepackage{multirow}

\begin{document}
\pagestyle{headings}
\mainmatter
\def\ECCVSubNumber{3586}  

\title{Auto-regressive Image Synthesis with \\ Integrated Quantization} 
\titlerunning{Auto-regressive Image Synthesis}

\author{Fangneng Zhan\inst{1,2}
\and
Yingchen Yu\inst{1} \and
Rongliang Wu\inst{1} \and
Jiahui Zhang\inst{1} \and
Kaiwen Cui\inst{1} \and
Changgong Zhang\inst{3} \and
Shijian Lu\inst{1}\thanks{Corresponding author}
}

\authorrunning{F. Zhan et al.}
\institute{Nanyang Technological University, Singapore 
\and
Max Planck Institute for Informatics, Germany
\and 
Amazon \\
\email{fzhan@mpi-inf.mpg.de}, \email{shijian.lu@ntu.edu.sg}, \email{cgzhang@amazon.com} \\
\email{\{yingchen001,ronglian001,jiahui003,kaiwen001\}@e.ntu.edu.sg} 
}

\maketitle

\begin{abstract} 
Deep generative models have achieved conspicuous progress in realistic image synthesis with multifarious conditional inputs, while generating diverse yet high-fidelity images remains a grand challenge in conditional image generation. This paper presents a versatile framework for conditional image generation which incorporates the inductive bias of CNNs and powerful sequence modeling of auto-regression that naturally leads to diverse image generation. Instead of independently quantizing the features of multiple domains as in prior research, we design an integrated quantization scheme with a variational regularizer that mingles the feature discretization in multiple domains, and markedly boosts the auto-regressive modeling performance. Notably, the variational regularizer enables to regularize feature distributions in incomparable latent spaces by penalizing the intra-domain variations of distributions. In addition, we design a Gumbel sampling strategy that allows to incorporate distribution uncertainty into the auto-regressive training procedure. The Gumbel sampling substantially mitigates the exposure bias that often incurs misalignment between the training and inference stages and severely impairs the inference performance. Extensive experiments over multiple conditional image generation tasks show that our method achieves superior diverse image generation performance qualitatively and quantitatively as compared with the state-of-the-art.
\end{abstract}

\section{Introduction}
\label{intro}

Conditional image generation aims to generate photorealistic images conditioning on certain guidance which can be semantic segmentation \cite{park2019spade}, key points \cite{tang2019cycle}, layout \cite{li2020bachgan} as well as heterogeneous guidance such as text \cite{ramesh2021zero} and audio \cite{chen2017deep}. It has been widely formulated as one-to-one mapping tasks~\cite{wang2018pix2pixhd}, though it is essentially one-to-many mappings since one conditional input could correspond to multiple images.
Targeting to mimic the true conditional image distribution, diverse yet high-fidelity image synthesis remains a great challenge in conditional image generation, especially when the conditional inputs come from different visual domains or even heterogeneous domains.

A typical approach to model diverse mapping is to employ extra style exemplars to guide the generation process.
For example, \cite{zhang2020cocosnet} build dense correspondences between conditional inputs and style exemplars to transfer textures for diverse generation, while building semantic correspondences essentially requires the exemplars to have similar semantics as the conditional inputs.
Without requiring extra exemplars, Variational Autoencoders (VAEs)~\cite{doersch2016tutorial} aim to regularize the latent distribution of encoded features, thus diverse generation can be achieved by directly sampling from the latent distribution. However, VAEs inevitably suffer from \textit{posterior collapse} phenomenon~\cite{lucas2019don} which leads to degraded diverse generation performance.
Instead of regularizing the latent feature distribution in VAE, VQ-VAE~\cite{oord2017neural} is designed to auto-regressively model the distributions of image feature sequences.
\cite{esser2020taming} further introduce transformers in VQ-VAE to achieve high-resolution image synthesis. Nevertheless, above auto-regressive generation methods discretize relevant features independently, neglecting the potential association among multi-domain features in latent spaces.

This paper presents an 
\textbf{I}ntegrated \textbf{Q}uantization \textbf{V}ariational \textbf{A}uto-\textbf{E}ncoder (\textbf{IQ-VAE})
that inherits the merits of CNNs (locality and spatial invariance) for high-fidelity image generation and the powerful sequence modeling of auto-regressive transformer for diverse image generation.
Instead of quantizing multi-domain features independently as in \cite{esser2020taming}, 
we introduce an integrated quantization scheme to quantize the involved features collaboratively in the latent spaces.
The integrated quantization scheme provides a sound way to regularize the latent structure of multi-domain distributions,
which can facilitate the ensuing auto-regressive modeling of sequence distributions.
However, as the conditional inputs and real images often have heterogeneous features with incomparable latent spaces, KL-divergence or Wasserstein distance cannot directly measure their feature discrepancy for regularization. 
Inspired by the differential circuit which takes the variation between two signals as valid input,
we introduce a variational regularizer which penalizes the intra-domain variation between distributions to regularize their structural discrepancy.

In addition, most auto-regressive models are trained with a so-called “teacher forcing” framework where the ground truth of target sequence (i.e., gold sequence) is provided at the training stage. However, such framework is susceptible to exposure bias, i.e., the misalignment between the training stage and the inference stage where the gold target sequence is not available and decisions are conditioned on previous model prediction.
We design a Gumbel sampling strategy that greatly mitigates the exposure bias by incorporating the uncertainty of sequence distributions in training stage. Specifically, we adopt a reparameterization trick with Gumbel softmax to samples tokens from the predicted distributions and then mixes them with the gold sequence according to a reliability-based scheduling to make the final prediction.
The Gumbel sampling also serves as data augmentation strategy that helps to avoid overfitting and improve the auto-regression performance substantially.

The contributions of this work can be summarized in three aspects. 
First, we introduce a versatile auto-regression framework with an integrated quantization scheme for conditional image generation.
Second, we propose a variational regularizer that exploits intra-domain variations to regularize heterogeneous features in latent spaces.
Third, we design a Gumbel sampling strategy with a reliability-based scheduling to mitigate the misalignment between the training and inference stages of auto-regressive models.

\section{Related Work}
\label{related}

\subsection{Conditional Image Generation}

Conditional image generation has achieved remarkable progress by learning the mapping among data of different domains. To achieve high-fidelity yet flexible image generation, various conditional inputs have been adopted including semantic segmentation \cite{isola2017pix2pix,wang2018pix2pixhd,park2019spade,zhan2022marginal,zhan2022modulated}, scene layouts \cite{sun2019lostgan,zhao2019layout2im,li2020bachgan}, key points \cite{ma2017pose,men2020adgan,zhan2021emlight,zhan2021gmlight}, edge maps \cite{isola2017pix2pix,zhan2021unite,zhan2021bi}, etc. Recently, several studies explored to generate images with cross-modal guidance \cite{zhan2021multimodal,yu2022towards}. 
For example, Qiao~\emph{et al.}~\cite{qiao2019mirrorgan} propose a novel global-local attentive and semantic-preserving text-to-image-to-text framework based on the idea of redescription.
Ramesh~\emph{et al.}~\cite{ramesh2021zero} handle text-to-image generation by using a transformer that auto-regressively models the text and image tokens.
Chen~\emph{et al.}~\cite{chen2017deep} investigated audio-to-visual generation with a conditional GANs. Nevertheless, the aforementioned methods all focus on deterministic image generation with a single generated image.

As an ill-posed problem, conditional image generation is a naturally a one-to-many mapping task as one conditional input could map to multiple diverse and faithful images. Earlier studies~\cite{kingma2013vae} manipulate latent feature codes to control the generation outcome, but they struggle to capture complex textures. With the emergence of GANs \cite{goodfellow2014generative,zhu2017unpaired,park2020contrastive,zhan2021spatial,zhan2019gadan}, style code injection has been designed to address this issue. For example, Zhu~\emph{et al.}~\cite{zhu2020sean} design semantic region-adaptive normalization (SEAN) to control the style of each semantic region individually. 
Choi~\emph{et al.}~\cite{choi2020starganv2} employ a style encoder for style consistency between exemplars and the translated images. 
Huang~\emph{et al.}~\cite{huang2018multimodal} and Ma~\emph{et al.}~\cite{ma2018exemplar} transfer style codes from exemplars to source images via adaptive instance normalization (AdaIN) \cite{huang2017adain}.
Recently, 
Zhang~\emph{et al.}~\cite{zhang2020cocosnet} learn dense semantic correspondences between conditional inputs and exemplars, but require the exemplars to have similar semantics with the conditional input.

The aforementioned methods all suffer from low performance in diverse generation or require extra guidance for decent diverse generation. In this work, we propose a versatile auto-regressive framework that introduces a joint quantization scheme to achieve conditional image generation, and it inherently allows to generate diverse yet high-fidelity images as well.

\subsection{Auto-regression in Image Generation}

Different from VAE or GANs in image generation, auto-regressive models treat image pixels as a sequence and generate pixels one by one conditioning on the previously generated pixels by modeling their conditional distributions. With the recent advance of deep learning, a number of studies explored to use deep auto-regressive models to generate image pixels sequentially. For instance, PixelRNN and PixelCNN~\cite{van2016pixel} utilize LSTM~\cite{hochreiter1997long} layers and masked convolutions to capture pixel inter-dependencies in a fixed order. 
Gated PixelCNN~\cite{oord2016conditional} describes a gated convolution to improve the generation quality with lower computational cost. However, deep auto-regressive models still struggle to generate high-fidelity images due to the limitation of sequential prediction of pixels. To address this issue, VQ-VAE~\cite{oord2017neural} adapts an encoder-decoder structure to learns discrete latent representations for autoregressive modeling, which enables high fidelity image synthesis.

Leveraging their powerful attention mechanisms, transformers~\cite{vaswani2017attention} allow to establish long-range dependencies  effectively and have been adopted in various computer vision tasks. In image generation, Chen~\emph{et al.}~\cite{chen2020generative} introduce a sequence Transformer to generate low-resolution images auto-regressively. Based on VQ-VAE~\cite{oord2017neural}, Esser~\emph{et al.}~\cite{esser2020taming} propose a VQ-GAN to learn a discrete codebook and utilize the transformers to efficiently model sequence distributions for high-resolution images synthesis. 
Nevertheless, the aforementioned methods all neglect exposure bias which often introduces clear misalignment between the training and the inference. The proposed Gumbel sampling strategy introduces uncertainty in training stage which mitigates the misalignment greatly.

\section{Proposed Method}
\label{method}

\subsection{Overall Framework}
\begin{figure*}[t]
\centering
\includegraphics[width=1\linewidth]{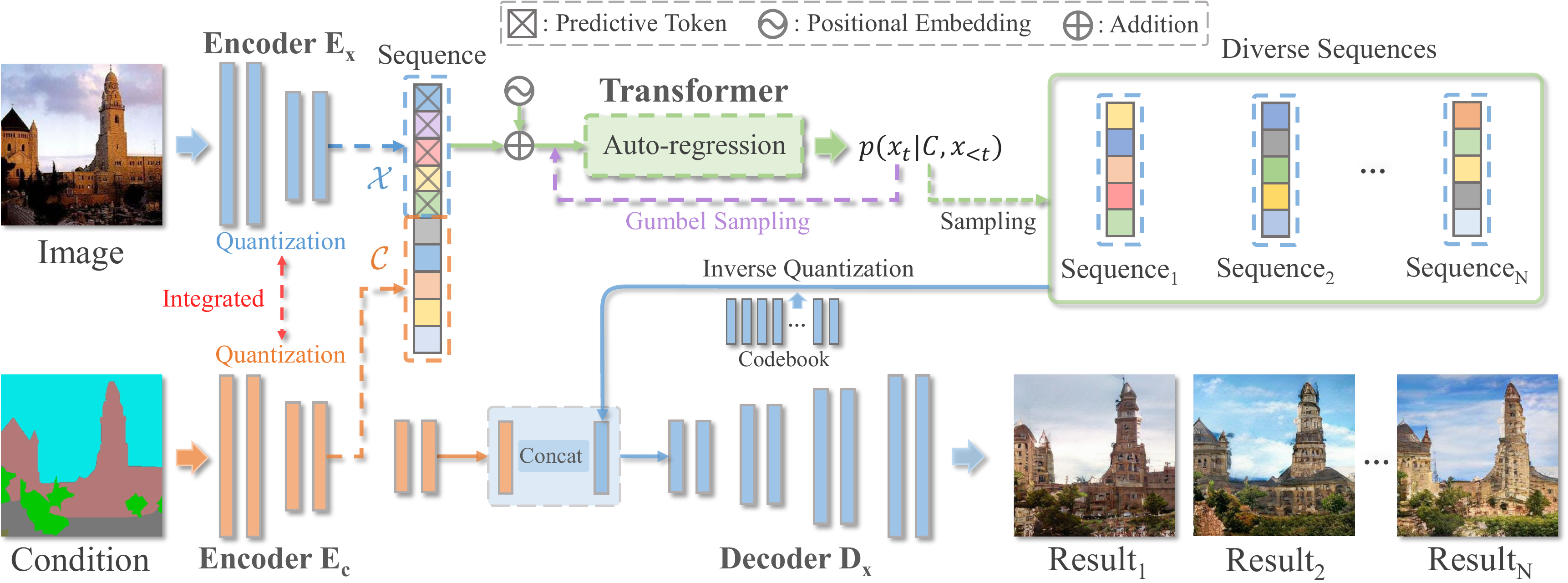}
\caption{
The framework of the proposed auto-regressive image generation with integrated quantization: We design an integrated quantization VAE (IQ-VAE) with $E_x$ and $E_c$ to encode the \textit{Image} and \textit{Condition} into discrete representation sequences $\mathcal{X}$ and $\mathcal{C}$ concurrently.
The distribution $p(x_t|\mathcal{C},x_{<t})$ of sequence $\mathcal{X}$ conditioned on $\mathcal{C}$ is modeled by an auto-regressive \textit{Transformer}. Finally, diverse sequences are sampled from the predicted distribution $p(x_t|\mathcal{C},x_{<t})$ which are further inversely quantized and concatenated with the encoded condition features for diverse generation via the IQ-VAE decoder $\rm D_{x}$.
}
\label{im_stru}
\end{figure*}

The framework of the proposed IQ-VAE is illustrated in Fig.~\ref{im_stru}.
The IQ-VAE is first trained to learn discrete feature representations of the real image and conditional input with learnable codebook as shown in Fig. \ref{im_joint} (a). With the learnt IQ-VAE and codebook, the conditional input and real image can be quantized into discrete sequences by IQ-VAE encoders $E_x$ and $E_c$. The transformer then auto-regressively models the distribution of the image sequences with a given sequence of conditional input. With the sequence distributions predicted by the transformer, diverse sequences can be sampled and inversely quantized into feature vectors based on the learnt codebook.
Finally, the inversely quantized feature vectors are concatenated with the conditional features and fed into the IQ-VAE decoder $D_{x}$ to achieve diverse image generation. Details of IQ-VAE and auto-regressive transformer will be discussed in the ensuing subsections.

\subsection{Integrated Quantization}
\label{method_iq}

For the task of conditional image generation, \cite{esser2020taming} employ two VQ-VAEs~\cite{oord2017neural} to quantize the features of conditional inputs and real images independently.
However, this naive quantization approach neglects the potential coupling between conditional inputs and real images in the latent spaces.
Intuitively, as conditional inputs imply certain information (e.g., edges) of the corresponding images, certain coupling or correlation should exist between their latent feature spaces.
Explicitly regularizing such coupling between images and conditional inputs will be beneficial for the modeling of image distribution from the given conditional inputs.

\begin{figure*}[t]
\centering
\includegraphics[width=1\linewidth]{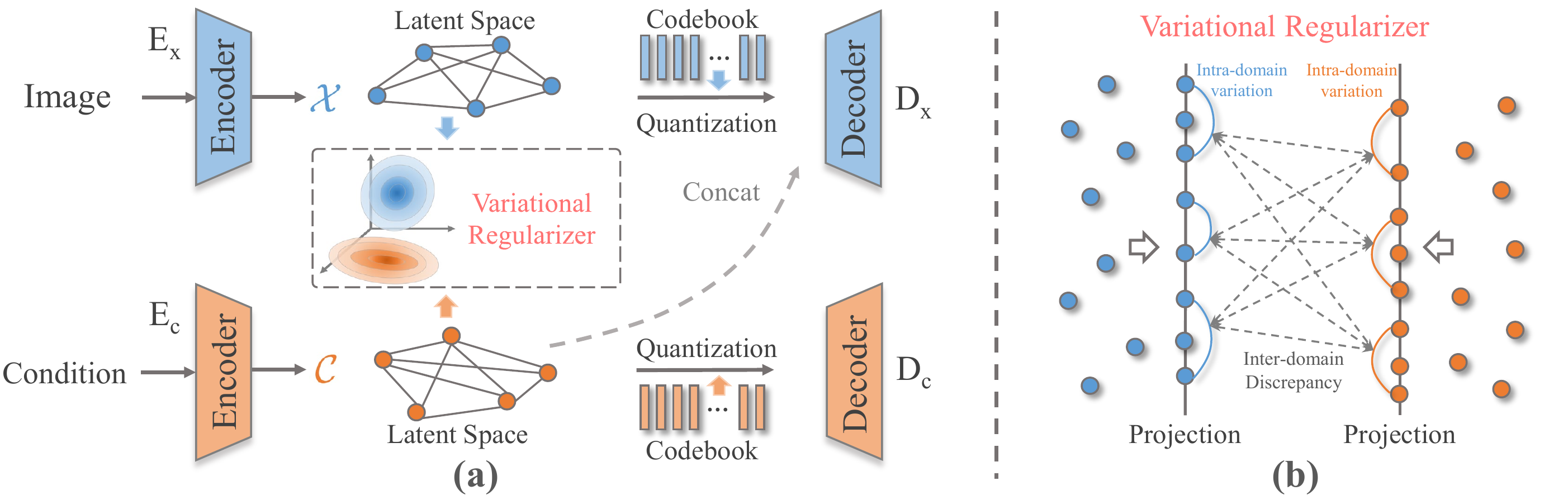}
\caption{
(a) illustrates the framework of the proposed integrated quantization scheme.
We introduce a variational regularizer to regularize their feature distributions in latent spaces.
As shown in (b), the variational regularizer employs the intra-domain variations to penalize the structural inter-domain discrepancy, and it is optimized through a sliced projection.
}
\label{im_joint}
\end{figure*}

We propose an integrated quantization scheme to regularize the discretization of the image and conditional input as illustrated in Fig.~\ref{im_joint}~(a).
Specially, two VQ-VAEs are employed to encode the image and conditional input to a pair of feature distributions as denoted by $\mathcal{X} = [x_1, x_2, \cdots, x_n]$ and $\mathcal{C} = [c_1, c_2, \cdots, c_n]$.
An intuitive method to regularize the feature distributions is to employ KL divergence to measure and minimize their inter-domain discrepancy, namely $\rm KL (\mathcal{C}, \mathcal{X})$.
However, this approach fails when a meaningful cost across the distributions cannot be defined.
This is especially true for heterogeneous conditional inputs (e.g., texts and audios) that have incomparable latent spaces with respect to the image. Under such context, the KL divergence is ill-suited and inapplicable to capture the discrepancy between distributions.
We thus design a novel variational regularizer that leverages the intra-domain variations of distributions to adaptively regularize their latent distributions.

\paragraph{Variational Regularizer}
Inspired by the differential circuit which takes the variation of two signals as the valid input,
we propose a variational regularizer that penalizes the inter-domain discrepancy via the intra-domain variations as illustrated in Fig.~\ref{im_joint}~(b).
Although the discrepancy between incomparable domain features $\mathcal{C} = [c_1, c_2, \cdots, c_n]$ and $\mathcal{X} = [x_1, x_2, \cdots, x_n]$ cannot be duly measured, the distance (or variation) among samples in the same domain can be effectively measured with some simple metric $\mathcal{M}$ (Euclidean distances is adopted in this work). We thus first compute the distances among intra-domain samples for the conditioned input and real image as denoted by $\mathcal{M}(c_i, c_k)$ and $\mathcal{M}(x_j, x_l)$. The discrepancy between intra-domain variations $\{\mathcal{M}(c_i, c_k)\}, i, k \in [1, n]$ and $\{\mathcal{M}(x_j, x_l)\}, j, l \in [1, n]$ can then serve as a proxy to indicate the inter-domain discrepancy between the conditional input and real image.

To regularize the structural difference between two latent distributions effectively, we adopt the discrete optimal transport (OT)~\cite{peyre2019computational,solomon2018optimal} with a $2^{th}$ Euclidean distance cost as the discrepancy metric which naturally induces the intrinsic geometries of distributions and can measure the discrepancy between intra-domain variations as follows:
\begin{equation}
\label{ot}
{\rm OT} (\mathcal{C}, \mathcal{X}) = \mathop{\min}\limits_{\Gamma \in \prod (\alpha, \beta)}  \sum_{i,j,k,l} \big| \mathcal{M}(c_{i}, c_{k}) - \mathcal{M} (x_{j}, x_{l}) \big|^{2} \Gamma_{ij}\Gamma_{kl} \\
\end{equation} 
where $\Gamma_{ij}$ and $\Gamma_{kl}$ are entries of coupling matrice $\Gamma$, 
$\prod (\alpha, \beta) = \{\Gamma \in \mathbb{R}^{n\times n} | \Gamma \Vec{1}_{n} = \alpha, \ \Gamma^{T} \Vec{1}_{n} = \beta \}$, 
$\Vec{1}_n$ is a n-dimensional all-one vector,
$\alpha=\{ \alpha_{i} \}$ and $\beta=\{ \beta_{j} \}, i,j \in [1, n]$ are vectors of probability weights associated with ${c_i}$ and ${x_j}$ ($\alpha_{i} = 1/n$, $\beta_{j} = 1/n$).
The formulation in Eq.~(\ref{ot}) is often referred as Gromov Wasserstein (GW) distance~\cite{memoli2011gromov} between distributions $\mathcal{C}$ and $\mathcal{X}$.

With GW distance as the metric in variational regularizer, we impose a constraint on the posterior distributions defined in different latent spaces which encourages structural similarity between them~\cite{xu2020learning}. This regularizer helps avoid over-regularization as it does not enforce a shared latent distribution across different or heterogeneous domains. In addition, the GW distance is invariant to translations, permutations or rotations on both distributions when Euclidean distances are used, which allows to capture discrepancy between complex latent distributions effectively.

\textbf{Optimization.}
The solution of the variational regularizer in Eq.~(\ref{ot}) is a non-convex optimization problem.
Grounded in the well-studied theory of Wasserstein disance~\cite{vayer2019sliced}, Eq.~(\ref{ot}) can be solved through sliced Gromov Wasserstein (sliced GW) distance~\cite{vayer2019sliced}.
Specifically, the original metric measure spaces are projected to 1D spaces with random directions, and the sliced GW corresponds to the expectation of the GW distances in these projected 1D spaces.
In this case, the sliced GW is approximated based on sample observations from the distributions shown in Fig.~\ref{im_joint}~(b).

In particular, given $[c_1, c_2, \cdots, c_n]$ from $\mathcal{C}$ and $[x_1, x_2, \cdots, x_n]$ from $\mathcal{X}$ and $L$ projection vectors $\{\gamma_m\}_{m=1}^{L}$, the empirical sliced GW can be formulated by:
\begin{equation}
\frac{1}{L} \sum_{m=1}^{L} \mathop{\min}\limits_{\Gamma_{ij}, \Gamma_{kl} \in \prod (p, q)}  \sum_{i,j,k,l} \big| \mathcal{M}(\langle c_i, \gamma_m \rangle , \langle c_j, \gamma_m \rangle) - \mathcal{M} (\langle x_k, \gamma_m \rangle , \langle x_l, \gamma_ m \rangle )
\big|^{2} \Gamma_{ij}\Gamma_{kl}.
\end{equation}
where $\langle c_i, \gamma_m \rangle$ denotes the projection of $c_i$ on direction $\gamma_{m}$.
Compared with direct computation via proximal gradient optimization \cite{xu2019gromov}, the sliced GW has much lower computational complexity of $\mathcal{O}(nd)$, where $n$ and $d$ denote the sample number and sample dimension, respectively.

Besides the loss of variational regularizer  (namely sliced GW) as denoted by $\mathcal{L}_{reg}$ for the optimization of IQ-VAE, we also include reconstruction loss $\mathcal{L}_{recon}$ and quantization loss $\mathcal{L}_{quan}$ of the conditional input and real image.
To further improve the image quality, a perceptual loss $\mathcal{L}_{perc}$ and discriminator loss $\mathcal{L}_{dis}$ are also included.
Thus, the overall objective for the IQ-VAE network is:  
\begin{equation}
    \mathcal{L}_{\scriptscriptstyle IQ-VAE} = \lambda_1 \mathcal{L}_{reg} + \lambda_2 \mathcal{L}_{recon} + \lambda_3 \mathcal{L}_{quan} + \lambda_4 \mathcal{L}_{perc} + \lambda_5 \mathcal{L}_{dis}.
\end{equation}
where $\lambda$ balances the loss terms.

\begin{figure*}[t]
\centering
\includegraphics[width=1\linewidth]{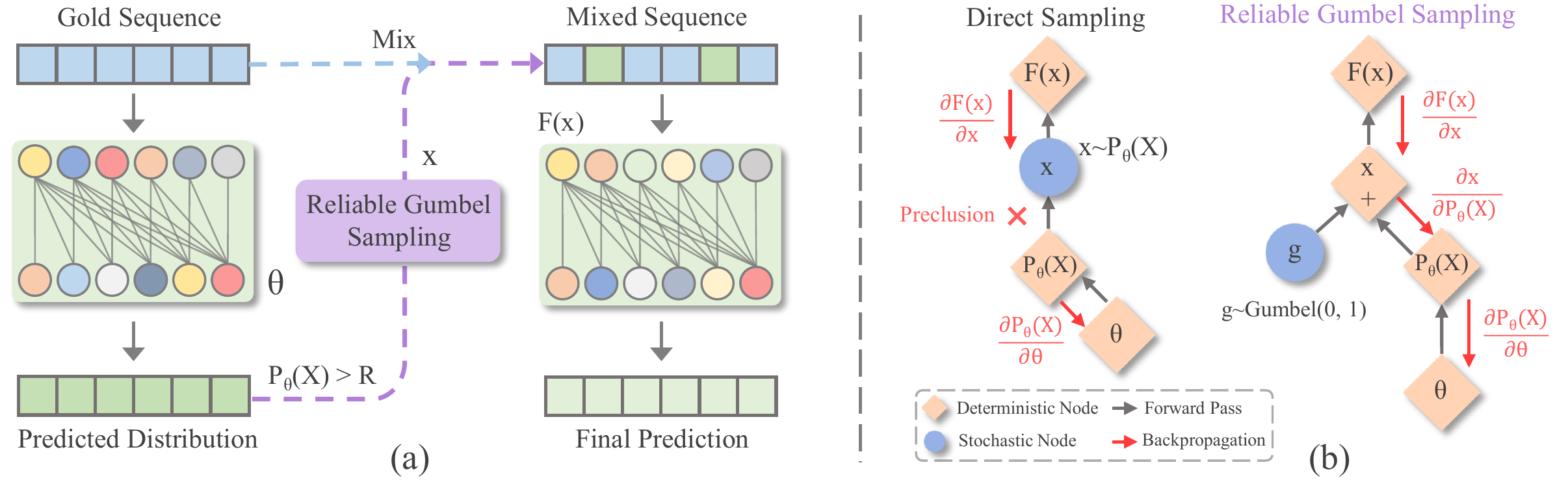}
\caption{
(a) illustrates the framework of the proposed Gumbel sampling with twice executions. In the first forward pass, token distribution $P_{\theta}(X)$ is predicted from the gold sequence with network parameters $\theta$. A sample $x$ is sampled from $P_{\theta}(X)$ according to a reliability-based scheduling and is mixed with the gold sequence for the second pass (namely final pass). (b) compares the gradient flows of direct sampling and Gumbel sampling. The presence of stochastic node $x$ in direct sampling precludes the backpropagation of gradient from $x$ to $P_{\theta}(X)$. Gumbel sampling allows gradient flow from $x$ to $P_{\theta}(X)$ through a reparameterization trick which transfers the stochasticity to a Gumbel distribution.
}
\label{im_gumbel}
\end{figure*}

\subsection{Auto-Regression}
\label{method_ar}

Auto-regressive (AR) modeling is representative objective to accommodate sequence dependencies in a raster scan order.
The probability of each position in the sequence is conditioned on all previously prediction and the joint distribution of sequences is modeled as the product of conditional distributions:
$
p(x) = \prod_{t=1}^{n} p(x_{t} | x_1, x_2, \cdots, x_{t-1}) = \prod_{t=1}^{n}p(x_t| x_{<t})
$.
Under the context of conditional image generation, a conditional auto-regression is actually adopted for the modeling of image distribution.
For clarity, we still denote the discrete image sequence as $\mathcal{X} = [ x_1, x_2,\dots, x_n ] $, the conditional sequence as $\mathcal{C} = [ c_1, c_2, \dots, c_n ] $. Then the joint distribution of image sequence conditioned on $\mathcal{C}$ can be formulated as:
\begin{equation}
p(x|\mathcal{C}) = \prod_{t=1}^{n} p(x_{t} | c_1, c_2, \cdots, c_n, x_1, x_2, \cdots, x_{t-1}) = \prod_{t=1}^{n}p(x_t| \mathcal{C}, x_{<t}).
\end{equation}

Auto-regressive models factorize the predicted tokens with chain rule of probability, which establishes the output dependency effectively for yielding better predictions.
During inference, each token is predicted auto-regressively in a raster-scan order. A top-$k$ ($k$ is 100 in this work) sampling strategy is adopted to randomly sample from the $k$ most likely next tokens, which naturally enables diverse sampling results.
The predicted tokens are then concatenated with the previous sequence as conditions for the prediction of next token. This process repeats iteratively until all the tokens are sampled.

\textbf{Gumbel sampling.}
Auto-regressive models are trained using the ground truth sequence (i.e., gold sequence).
This framework leads to quick convergence during training, but it is misaligned with the inference stage where gold sequence is not available and decisions are purely conditioned on previous predictions. This phenomenon is typically referred as exposure bias~\cite{schmidt2019generalization}.
Intuitively, this problem can be tackled by using the previous predictions as conditions with certain probability in training stage as mentioned in \cite{mihaylova2019scheduled}.

Specially, in order to conduct sampling from previous predictions, the auto-regression process is executed twice in training stage as illustrated in Fig.~\ref{im_gumbel}.
In the first execution, the predictions are conditioned on the gold sequence and yield discrete distribution $p_{\theta}(X)=[p_1, \cdots, p_l ]$  for each token ($\theta$ is network parameter, $l$ is the number of codebook embedding).
In the second execution, we aim to sample tokens according to the discrete distributions.
However, direct sampling from a distribution will preclude the gradient backpropagation as shown in Fig.~\ref{im_gumbel}~(b).
A Gumbel sampling strategy is thus introduced with a reparameterization trick \cite{jang2016categorical} to enable gradient backpropagation in discrete distribution sampling.
Specially, the sampling operation is conducted on a Gumbel-softmax distribution~\cite{jang2016categorical} which is defined by:
${\rm softmax}(1/\tau(p_{\theta}(X)+g))$,
where $g\sim \mathrm{Gumbel}(0,1) = -\log(-\log U)$, $U\sim \mathrm{Uniform}(0, 1)$.
A sample $x_i$ drawn from the Gumbel-softmax distribution can be denoted by:
\begin{align*}
x_i = \frac{\exp((\log(p_i)) + g_i)/ \tau)}
{\sum_{j=1}^{n} \exp((\log(p_j) + g_j) / \tau)} 
\quad {\rm for} \; i = 1,2, \cdots, n.
\end{align*} 
where $\tau$ is an annealing parameter.
The sampling from a Gumbel-softmax distribution exactly approximates the sampling from the categorical distribution $p_{\theta}(X)$ as proved in \cite{maddison2014sampling}.
In forward pass of network training, sampling is actually conducted on the Gumbel(0,1) distribution which is independent of the network parameter $\theta$.
In backpropagation, the sampling operation is not involved in the gradient flow, which means that the stochasticity of sampling operation is transferred from $p_{\theta}(X)$ to the Gumbel(0,1) distribution.

To schedule the sampling in accordance with the training process, we design a \textit{Gumbel sampling} strategy based on the prediction reliability.
Considering sampled tokens are more difficult to learn than the ground truth especially at the early training stage,
we only sample tokens for positions with high prediction reliability as denoted by $R$ \cite{liu2021confidence}.
For a ground truth embedding $\gamma_{i}$ and predicted distributions $[p_1, \cdots, p_l ]$ associated with normalized codebook embeddings $[\gamma_{1}, \dots, \gamma_{l}]$, the prediction reliability $R_i$ can be quantified by the weighted summation of the inner products of embeddings:
\begin{equation}
    R_{i} = 
    \sum_{j=1}^{l}p_j*\gamma_{j}\cdot \gamma_i  \ , \quad i \in [1, n] \ .
\end{equation}
$R_i \in [0, 1]$ accurately indicates the similarity between the predicted token distribution and the ground truth token, and measures whether the prediction reliability reaches the threshold (0.9 by default) to conduct token sampling.

After obtaining a sequence representing the model prediction for each position, 
we mix the gold tokens and predicted tokens with a given probability which is a function of the training step and is calculated with a selected schedule. 
We then pass the new mixed sequence to the transformer for the second execution to yield the final predictions.
Note that only the gradient of the second execution is backpropagated in model training.

\textbf{Computational cost.}
Twice execution for Gumbel sampling will increase the training time, which can be mitigated by reducing the frequency of applying Gumbel sampling.
In our implementation, the Gumbel sampling is applied for every 4 iterations by default.
The average speed of our model with Gumbel sampling is 2.8 iteration/s, and the model speed without Gumbel sampling is 3.0 iteration/s.
Therefore, the increase of computational cost is very limited.

\renewcommand\arraystretch{1.1}
\begin{table*}[t]
\scriptsize
\caption{
Comparing IQ-VAE with state-of-the-art image generation methods over four conditional image generation tasks. The adopted evaluation metrics include FID, SWD and LPIPS.
}
\renewcommand\tabcolsep{2pt}
\centering 
\begin{tabular}{l||ccc||ccc||ccc} \hline
\multirow{2}{*}{\textbf{Methods}}
& 
\multicolumn{3}{c||}{\textbf{ADE20K}} & 
\multicolumn{3}{c||}{\textbf{CelebA-HQ(Edge)}} &
\multicolumn{3}{c}{\textbf{DeepFashion}}
\\
\cline{2-10}
& FID $\downarrow$ & SWD $\downarrow$ & LPIPS $\uparrow$ & FID
$\downarrow$ & SWD  $\downarrow$ & LPIPS  $\uparrow$ & FID  $\downarrow$ & SWD  $\downarrow$ & LPIPS  $\uparrow$ \\\hline 

\textbf{Pix2pixHD} \cite{wang2018pix2pixhd} & 61.08 & 28.47 & N/A         & 42.70 & 33.30 & N/A     & 25.20  & \textbf{16.40}  & N/A           \\

\textbf{Pix2pixSC} \cite{wang2019pix2pixsc} & 56.23 & 24.52 & 0.378        & 49.39 & 33.20 & 0.193     & 28.49 & 21.13  & 0.172           \\

\textbf{BicycleGAN} \cite{zhu2017toward} & 62.52 & 33.27 & 0.405           & 44.63 & 31.96 & 0.224       & 29.82  & 22.74  & 0.251  \\

\textbf{StarGAN~v2} \cite{choi2020starganv2} & 98.72 & 65.47 & \textbf{0.451}         & 48.63 & 41.96 & 0.214     & 43.29  & 30.87  & 0.296           \\

\textbf{DRIT++} \cite{lee2020drit++}     & 105.1 & 81.82 & 0.432          & 50.31  & 47.21  & 0.313         & 52.67 & 42.34 &  0.281    \\

\textbf{SPADE} \cite{park2019spade} & 33.90 & 19.70 & 0.344           & 31.50 & 26.90 & 0.207        & 36.20  & 27.80  & 0.231         \\

\textbf{SMIS} \cite{zhu2020smis} & 42.17 &  22.67 &  0.416             &  23.71  &  22.23  & 0.201           & 26.23  & 23.73  & 0.240           \\

\textbf{VQ-GAN} \cite{esser2020taming}     & 35.50 & 21.50 & 0.421             & 16.23 & 23.33 & 0.330        & 16.49  & 21.20 & 0.314           \\

\hline
\textbf{IQ-VAE}
& \textbf{29.77} & \textbf{17.44} & 0.447
& \textbf{14.71} & \textbf{19.74} & \textbf{0.344}
& \textbf{11.15} & 19.01 & \textbf{0.320}
  \\\hline
\end{tabular}

\label{tab_com}
\end{table*}

\begin{figure*}[t]
\centering
\includegraphics[width=1\linewidth]{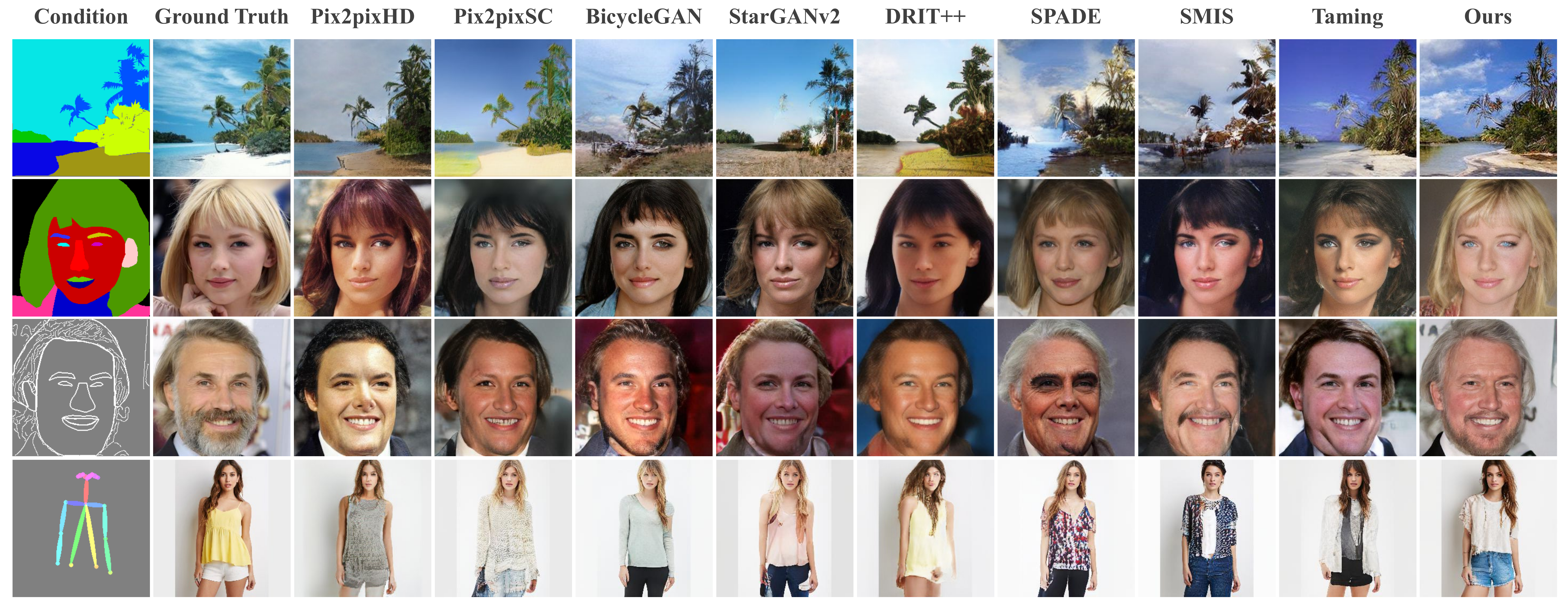}
\caption{
Qualitative illustration of IQ-VAE and state-of-the-art image generation methods over four types of generation tasks. 
IQ-VAE is able to generate faithful images with high fidelity.
}
\label{im_com}
\end{figure*}

\section{Experiments}
\label{experiment}

\subsection{Experimental Settings}
\label{setting}

\textbf{Datasets.}
We benchmark our method over multiple public datasets in conditional image generation.

\noindent
$\bullet$ ADE20k \cite{zhou2017ade20k} has 20k training images associated with a 150-class segmentation mask. We use its semantic segmentation as conditional inputs in experiments.

\noindent
$\bullet$ CelebA-HQ \cite{liu2015celebahq} has 30,000 high quality face images whose semantic maps and edges serve as the condition for image generation. 

\noindent
$\bullet$ DeepFashion \cite{liu2016deepfashion} has 52,712 person images of different appearances and poses.
We use the key points of the person images as conditional inputs in experiments.

\noindent
$\bullet$ COCO-Stuff \cite{caesar2018cocostuff} augments COCO \cite{lin2014coco} with pixel-level stuff annotations. We use its layout as condition for image generation. 

\noindent
$\bullet$ CUB-200 \cite{welinder2010caltech}  has 200 bird species with attribute labels and we use it for text-to-image generation.

\noindent
$\bullet$ Sub-URMP \cite{chen2017deep} is a subset of URMP \cite{li2018creating} and we use it for audio-to-image generation.

\noindent
\textbf{Evaluation Metrics.} 
We evaluate the proposed IQ-VAE on the tasks of semantic-to-image, edge-to-image and keypoint-to-image generation, as these tasks have rich prior studies for comprehensive yet fair benchmarking.
We assess the compared methods with several widely adopted evaluation metrics. Specifically, \textit{Fr{\'e}chet Inception Score (FID)}~\cite{fid} and \textit{Sliced Wasserstein distance (SWD)}~\cite{swd} are employed to evaluate the quality of generated images. \textit{Learned Perceptual Image Patch Similarity (LPIPS)}~\cite{zhang2018lpips} measures the distance between image patches, which is employed to evaluate the diversity of generated images and reconstruction performance of auto-encoder.

\noindent
\textbf{Implementation Details.}
The proposed model is optimized with a learning rate of 1.5$e$-4.
The auto-regressive transformer is implemented based on the GPT2 architecture~\cite{radford2019language} with a input size of $256$.
AdamW~\cite{loshchilov2017decoupled} solver is adopted with $\beta_{1}=0.9$ and $\beta_{2}=0.95$. All experiments are conducted on 4 Tesla V100 GPUs with a batch size of 32.
The size of generated images is $256 \times 256$ for all evaluated generation tasks.
The transformer is implemented based on minGPT~\footnote{https://github.com/karpathy/minGPT}.
Table. \ref{tab_hyper} shows parameter setting in the transformer and IQ-VAE.

\begin{table*}[t]
\centering
\renewcommand\tabcolsep{3pt}
\begin{tabular}{lc | lc} 
\multicolumn{2}{c|}
{Transformer } 
&
\multicolumn{2}{c}
{IQ-VAE}  \\\hline
Parameters & Setting &  Parameters & Setting
\\
learning rate & 1.5$e$-4   & learning rate & 1.5$e$-4  \\
batch size & 32      & batch size & 32  \\
epoch & 50           & epoch  &  100  \\
vocabulary size & 1024 &  codebook embedding number  & 1024   \\
embedding number  & 1024 & codebook embedding dimension  & 256 \\
sequence length & 512     &  feature number & 256 \\
number of transformer block & 24   &    &    \\
\end{tabular}
\caption{
The parameter setting in the proposed transformer and IQ-VAE.
}
\label{tab_hyper}
\end{table*}

\begin{figure*}[t]
\centering
\includegraphics[width=1\linewidth]{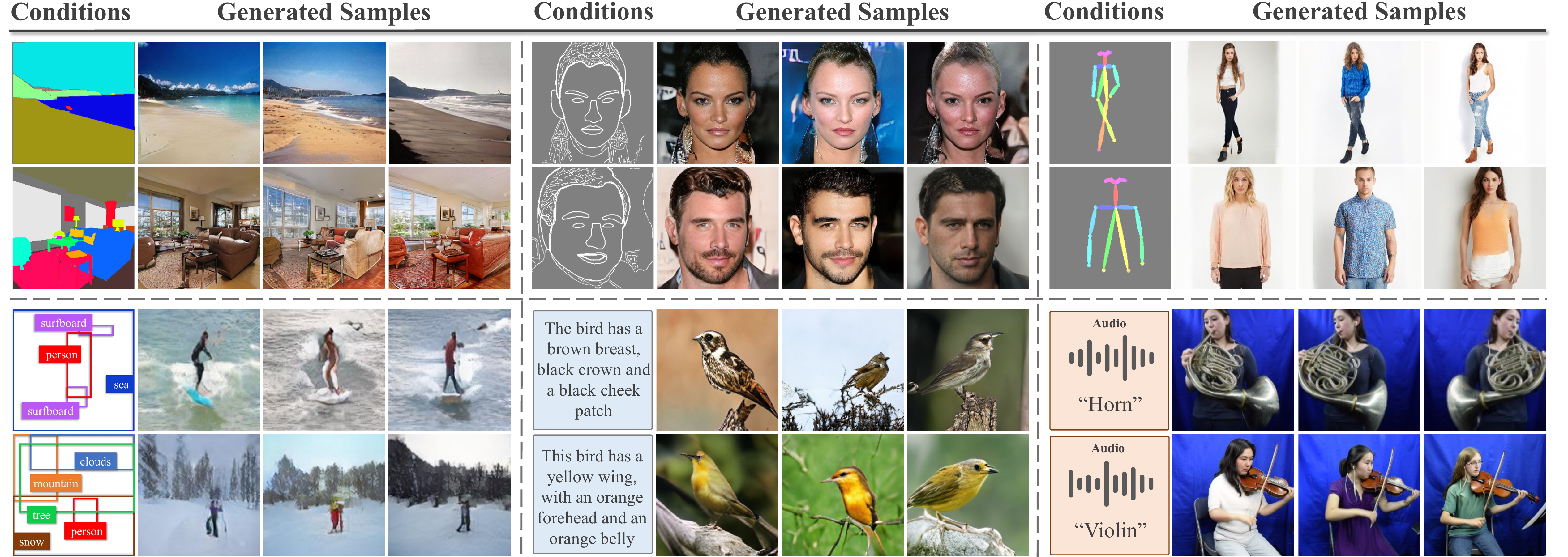}
\caption{
Illustration of diverse image generation by the proposed IQ-VAE: Faithful yet diverse images are successfully generated with different types of conditional inputs such as semantic maps, edge maps, key points, layout maps, as well as heterogeneous conditions like texts and audios.
}
\label{im_diverse}
\end{figure*}

\subsection{Quantitative Results} 
We compare the proposed IQ-VAE with several state-of-the-art conditional image generation methods including 
1) Pix2pixHD~\cite{wang2018pix2pixhd};
2) Pix2pixSC~\cite{wang2019pix2pixsc};
3) BicycleGAN~\cite{zhu2017toward};
4) StarGAN~v2~\cite{choi2020starganv2};
5) DRIT++~\cite{lee2020drit++};
6) SPADE~\cite{park2019spade};
7) SMIS~\cite{zhu2020smis};
8) Taming Transformer~\cite{esser2020taming}.

In the quantitative experiments, all compared methods generate diverse images except Pix2PixHD \cite{wang2018pix2pixhd} which does not support diverse generation. Table~\ref{tab_com} shows experimental results in FID, SWD and LPIPS. It can be observed that IQ-VAE outperforms all compared methods across most metrics and tasks consistently.
DRIT++~\cite{lee2020drit++} and StarGAN v2~\cite{choi2020starganv2} achieve relatively high LPIPS scores by sacrificing the image quality as measured by FID and SWD, while SPADE~\cite{park2019spade} and SMIS~\cite{zhu2020smis} achieve decent FID and SWD scores with degraded LPIPS scores. 
The proposed IQ-VAE employs powerful variational auto-encoders to achieve high-fidelity image synthesis and a auto-regressive model for faithful image diversity modeling, thus achieving superior performance in terms of image quality and diversity.
Compared with Taming transformer~\cite{esser2020taming}, the proposed IQ-VAE allows to quantize the image sequences and conditional sequence jointly and boosts the auto-regressive modeling for better FID and SWD scores.
In addition, the proposed Gumbel sampling introduces uncertainty of distribution sampling into the training process which mitigates the exposure bias and improves the inference performance clearly. As the mixed sequence serves as certain extra data augmentation, the Gumbel sampling also helps to alleviate the over-fitting of auto-regressive model effectively.

\renewcommand\arraystretch{1.05}
\begin{table}[t]
\small 
\caption{
Ablation study of IQ-VAE on ADE20k. VR and None denote the proposed variational regularizer and no regularization, respectively. GS denotes the proposed Gumbel sampling.}
\centering
\label{tab_ablation}
\renewcommand\tabcolsep{8.5pt}
\begin{tabular}
{l | c  c  c }
\hline 
\textbf{Models} & \textbf{FID}  $\downarrow$ & \textbf{SWD}  $\downarrow$ & \textbf{LPIPS}  $\uparrow$
\\
\hline\hline
\textbf{VQ-GAN}        & 35.50  & 21.50 &  0.421  \\
\textbf{IQ-VAE(None)}  & 31.88  & 19.14  & 0.441   \\
\textbf{IQ-VAE(VR)}    & 31.41  & 18.71 & \textbf{0.450}  \\
\hline
\textbf{IQ-VAE(VR) + GS}   & \textbf{29.77} & \textbf{17.44} &  0.447  \\
\hline
\end{tabular}

\end{table}

\subsection{Qualitative Evaluation} 
We perform qualitative comparisons as shown in Fig.~\ref{im_com}.
The experiments are conducted over six datasets including 
ADE20k~\cite{zhou2017ade20k},
CelebA-HQ~\cite{liu2015celebahq},
DeepFashion~\cite{liu2016deepfashion},
COCO-Stuff~\cite{caesar2018cocostuff},
CUB-200~\cite{welinder2010caltech},
and Sub-URMP~\cite{chen2017deep}.
The splits of training and testing sets on all above datasets follow the default split settings.
In addition, the data used in the experiments do not contains person identity related information or offensive contents.
It can be seen that IQ-VAE achieves the best visual quality and presents remarkable coherence with the condition.
SPADE~\cite{park2019spade} and SMIS~\cite{zhu2020smis} adopt VAE to constraint the distribution of encoded features which cannot capture the complex distributions of real images. StarGAN~v2~\cite{choi2020starganv2} and DRIT++~\cite{lee2020drit++} adopt single latent code to encode image styles, which tends to capture global styles but misses local details.

IQ-VAE also generalizes well and demonstrates superior synthesis quality and diversity in various generation tasks as illustrated in Fig. \ref{im_diverse}. It can be observed that IQ-VAE is capable of synthesizing high-fidelity images with various conditional inputs such as semantic maps, edge maps, keypoints, layout maps as well as heterogeneous conditions such as texts and audios.

\subsection{Ablation Study}
\label{ablation}

We conduct extensive ablation studies to evaluate IQ-VAE as shown in Table \ref{tab_ablation}.
The baseline is selected as VQ-GAN (namely Taming Transformer~\cite{esser2020taming}).
Replacing VQ-GAN with the proposed IQ-VAE without any regularization in IQ-VAE(None) brings in marginal improvement.
The proposed variational regularizer with adaptive weights in IQ-VAE(VR) improves the generation performance, demonstrating the effectiveness of adaptive weights learning.
Finally, including the Gumbel sampling remarkably boosts the performance as indicated in IQ-VAE(VR)+GS.

\begin{wrapfigure}{r}{0.5\textwidth}
\centering
\includegraphics[width=1.0\linewidth]{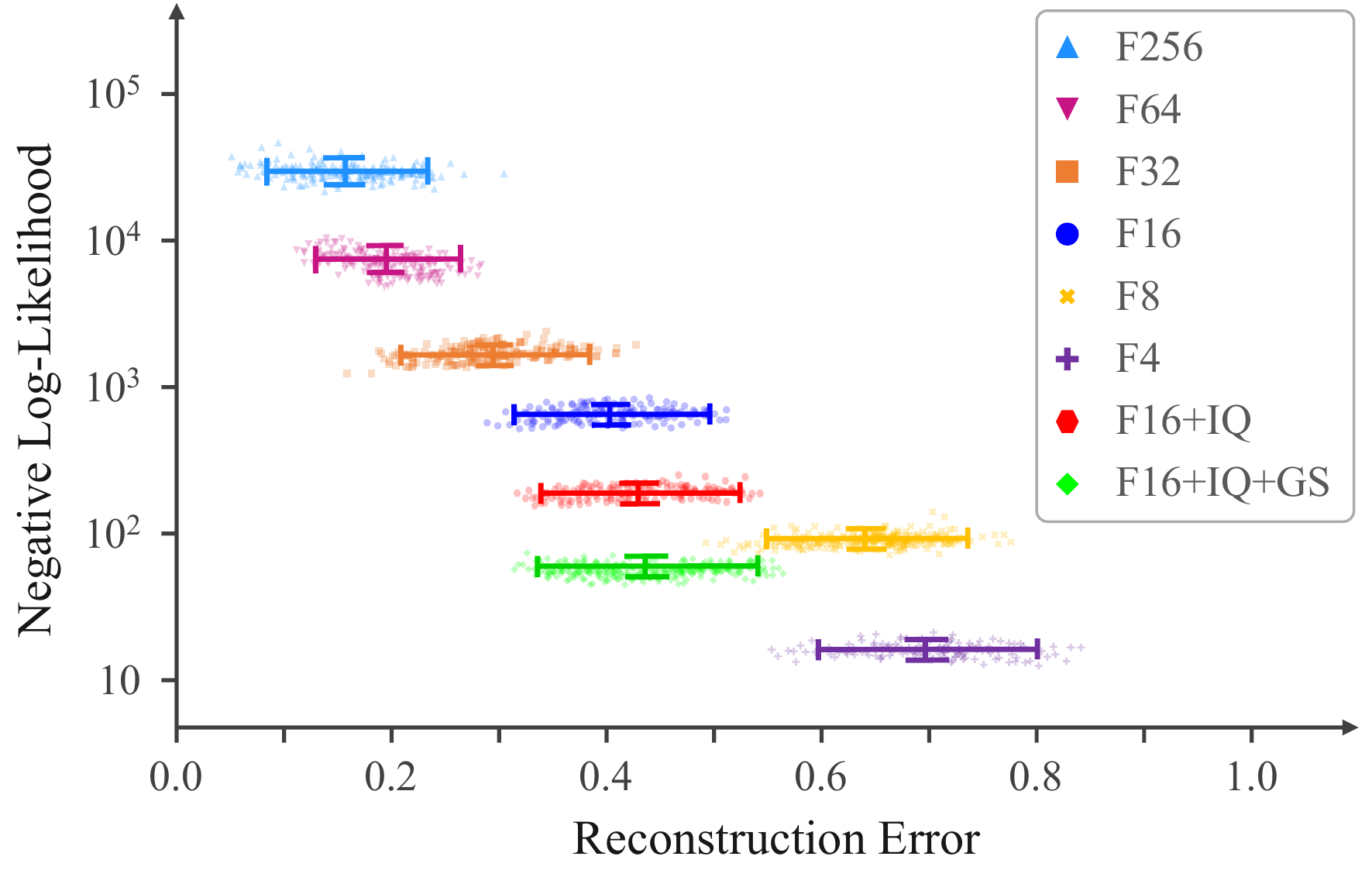}
\caption{
Trade-off between negative log-likelihood and reconstruction error with different sizes of encoded features on CelebaHQ~\cite{liu2015celebahq}.
}
\label{im_graph}
\end{wrapfigure}

We study the effect of feature sizes for discrete representation in IQ-VAE and Fig.~\ref{im_graph} shows experimental results on the CelebaHQ dataset. As Fig.~\ref{im_graph} shows, 
we specify the size of representation features in terms of a factor $F$ where $Fx$ denotes a feature size of $x\times x$.
Note the input size of transformer is always fixed at $16\times 16$. The horizontal axis of the graph shows reconstruction error as measured by LPIPS~\cite{zhang2018lpips} which indicates the upper bound of generation quality (lower is better), while the vertical axis shows negative log-likelihood from the transformer which indicates the performance of the auto-regressive modeling (lower is better). We can see that there is a trade-off between the negative log-likelihood and reconstruction error. Though an encoded feature of small size allows the transformer to better model the image distribution, the reconstruction deteriorates severely after a certain value (F16 in this case). The proposed integrated quantization and Gumbel sampling instead improve the negative log-likelihood remarkably without sacrificing the reconstruction performance clearly.

\subsection{User Study}
\label{user}

\begin{figure*}[ht]
\centering
\includegraphics[width=1.0\linewidth]{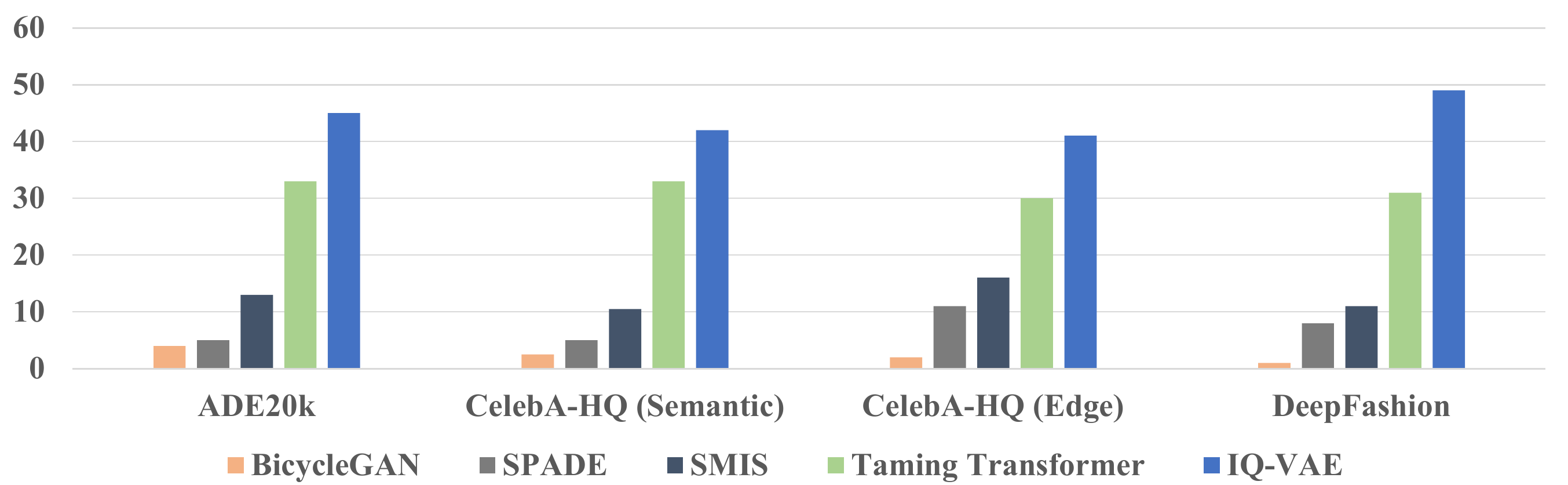}
\caption{
User study over four datasets ADE20K~\cite{zhou2017ade20k}, CelebA-HQ~\cite{liu2015celebahq}(Semantic), CelebA-HQ~\cite{liu2015celebahq}(Edge), DeepFashion \cite{liu2016deepfashion}.
The bars show the number of images that AMT users ranked with the best visual quality.
}
\label{im_amt}
\end{figure*}

We conduct crowdsourcing user study to evaluate the quality of generated images as shown in Fig.~\ref{im_amt}.
Specifically, 100 pairs of images generated by all compared methods are shown to 10 users who selected the image with the best visual quality. 
As shown in Fig.~\ref{im_amt}, we compared the proposed IQ-VAE with several state-of-the-art generation methods including BicycleGAN~\cite{zhu2017toward}, SPADE~\cite{park2019spade}, SMIS~\cite{zhu2020smis}, and Taming Transformer~\cite{esser2020taming}. The images generated by the proposed IQ-VAE are much more realistic according to the user feedback.

\section{Conclusions}
\label{conclusion}
This paper presents IQ-VAE, an auto-regressive framework with integrated quantization for conditional image synthesis.
We propose a novel variational regularizer to regularize the feature distribution structures of conditional inputs and real images, which boosts the auto-regressive modeling clearly.
To mitigate the misalignment between training and inference of auto-regressive model, a Gumbel sampling strategy with a reliability-based scheduling is included in the training stage and improves the inference performance by a large margin. 
Quantitative and qualitative experiments show that IQ-VAE is capable of generating diverse yet high-fidelity images with multifarious conditional inputs.

\textbf{Limitations.}
As auto-regression is adopted in the model to predict image sequence, the inference speed is inevitably constrained which may limit the application of the proposed model in time-critical tasks. 
Although some works \cite{wiggers2020predictive,oord2018parallel} have been proposed to speed up the autoregressive sampling, the acceleration for the inference of auto-regressive model is still an open challenge.

\textbf{Potential Negative Societal Impacts}
This work aims to synthesize diverse yet high-fidelity images with given conditional inputs. It could have negative impacts if it is used for certain illegal purpose such as image forgery and manipulation.

\textbf{Acknowledgement.}
This study is supported under the RIE2020 Industry Alignment Fund – Industry Collaboration Projects (IAF-ICP) Funding Initiative, as well as cash and in-kind contribution from the industry partner(s).

\clearpage
%
%
\bibliographystyle{splncs04}
\bibliography{egbib}
\end{document}